\documentclass[conference]{IEEEtran}
\usepackage{times}

\usepackage[numbers]{natbib}
\usepackage{multicol}
\usepackage{hyperref}

\usepackage{subfig}
\usepackage{float}
\usepackage[linesnumbered,ruled,vlined]{algorithm2e}
\usepackage{bbm}
\usepackage{siunitx}
\usepackage{amsmath,amssymb,amsfonts}
\usepackage{makecell}
\usepackage{graphicx}
\usepackage{xcolor}
\usepackage{bm}
\usepackage{todonotes}
\usepackage[belowskip=-2pt,aboveskip=3pt]{caption}
\definecolor{green}{RGB}{255,128,0}
\definecolor{red}{RGB}{200,20,20}

\SetAlCapSkip{0.5em}

\usepackage{bbm}
\usepackage{bm}
\usepackage{siunitx}
\usepackage{amsmath,amssymb,amsfonts}

\newcommand{\dataset}{\mathcal{D}}
\newcommand{\nStep}{N_p}
\newcommand{\mProj}{M_p}
\newcommand{\state}{s}
\newcommand{\robot}{r}

\newcommand{\env}{e}
\newcommand{\action}{a}
\newcommand{\classLabel}{w}
\newcommand{\transform}{T}
\newcommand{\aug}[1]{\Tilde{#1}}
\newcommand{\test}[1]{{#1}'}
\newcommand{\distfname}{\mathrm{dist}}
\newcommand{\distf}[2]{\distfname(#1,#2)}
\newcommand{\target}[1]{#1^\text{target}}
\newcommand{\example}{x}
\newcommand{\exampleSpace}{\mathcal{X}}
\newcommand{\relevantSet}{\mathcal{X}_r}
\newcommand{\validSet}{\mathcal{X}_v}
\newcommand{\augf}{\phi}
\newcommand{\augDef}{\augf:\exampleSpace\rightarrow\exampleSpace}

\newcommand{\exampleAug}{\aug{\example}}
\newcommand{\nAug}{k}

\newcommand{\exampleAugSet}{\exampleAug_{1:\nAug}}
\newcommand{\transformDist}{p_{\exampleAugSet}(\transform)}
\newcommand{\transformUniformRange}{[\transform^-,\transform^+]}
\newcommand{\transformUniform}{\mathbb{U}\transformUniformRange}
\newcommand{\loss}{\mathcal{L}}
\newcommand{\lossRobot}{\loss_\text{robot}}
\newcommand{\lossBbox}{\loss_\text{bbox}}
\newcommand{\lossValid}{\loss_\text{valid}}
\newcommand{\lossOcc}{\loss_\text{occ}}
\newcommand{\validF}{f_\text{valid}}
\newcommand{\learnedValidF}{f_{\text{valid},\theta}(\transform)}
\newcommand{\minDist}{{d^-}}
\newcommand{\betaBbox}{\beta_1}
\newcommand{\betaValid}{\beta_2}
\newcommand{\betaOcc}{\beta_3}
\newcommand{\betaDmd}{\beta_4}
\newcommand{\lossDmd}{\loss_{\Delta\minDist}}

\newcommand{\projectionNotProgressing}{\delta_p}

\newcommand{\projStepSizeThreshold}{\epsilon_p}
\newcommand{\points}{p}
\newcommand{\contactpoint}{p^c}
\newcommand{\statePoint}{\points_{\state,i}}
\newcommand{\augStatePoint}{\aug{\points}_{\state,i}}
\newcommand{\statePoints}{\points_\state}
\newcommand{\augStatePoints}{\aug{\points}_\state}
\newcommand{\pointMinDist}{p_\minDist}
\newcommand{\augPointMinDist}{\aug{p}_\minDist}

\newcommand{\SDF}{\mathrm{SDF}}

\newcommand{\learnValidError}{y_\text{valid}}
\newcommand{\minLearnValidError}{\learnValidError^-}
\newcommand{\learnValidDataset}{\mathcal{D}_\text{valid}}
\newcommand{\transformDim}{d}
\newcommand{\learnValidScaling}{\alpha_\text{valid}}
\newcommand{\nLearnValid}{n_\text{valid}}

\newcommand{\klF}{D_{KL}}
\newcommand{\kl}[2]{\klF({#1}\,||\,{#2})}
\newcommand{\uniformity}{e^{-\kl{\transformDist}{\transformUniform}}}
\newcommand{\apply}{\texttt{apply}}
\newcommand{\applyState}{\texttt{apply\_state}}
\newcommand{\sample}{\texttt{sample}}
\newcommand{\augState}{\texttt{aug\_state}}
\newcommand{\augRobot}{\texttt{aug\_robot}}
\newcommand{\learnValidSimF}{\texttt{simulate}}
\newcommand{\lossDiversity}{\loss_\mathbb{U}}
\newcommand{\learnValidStateActionSet}{Q_\text{valid}}

\begin{document}

\title{Data Augmentation for Manipulation}

\author{\authorblockN{Peter Mitrano}
\authorblockA{University of Michigan\\
Email: pmitrano@umich.edu}
\and
\authorblockN{Dmitry Berenson}
\authorblockA{University of Michigan\\
Email: dmitryb@umich.edu}
}

\maketitle

\begin{abstract}
The success of deep learning depends heavily on the availability of large datasets, but in robotic manipulation there are many learning problems for which such datasets do not exist. Collecting these datasets is time-consuming and expensive, and therefore learning from small datasets is an important open problem. Within computer vision, a common approach to a lack of data is \textit{data augmentation}. Data augmentation is the process of creating additional training examples by modifying existing ones. However, because the types of tasks and data differ, the methods used in computer vision cannot be easily adapted to manipulation. Therefore, we propose a data augmentation method for robotic manipulation. We argue that augmentations should be valid, relevant, and diverse. We use these principles to formalize augmentation as an optimization problem, with the objective function derived from physics and knowledge of the manipulation domain. This method applies rigid body transformations to trajectories of geometric state and action data. We test our method in two scenarios: 1) learning the dynamics of planar pushing of rigid cylinders, and 2) learning a constraint checker for rope manipulation. These two scenarios have different data and label types, yet in both scenarios, training on our augmented data significantly improves performance on downstream tasks. We also show how our augmentation method can be used on real-robot data to enable more data-efficient online learning.
\end{abstract}

\IEEEpeerreviewmaketitle

\section{Introduction}

In recent years, interest in applying deep learning to robotic manipulation has increased. However, the lack of cheap data has proven to be a significant limitation \cite{DisbandOpenAI2021}. To enable applications such as smart and flexible manufacturing, logistics, and care-giving robots \cite{WEFRobots}, we must develop methods that learn from smaller datasets, especially if the learning is done online on real robots.

One of the simplest and most effective ways to mitigate the problem of small datasets is to use data augmentation. While data augmentation has been shown to significantly improve generalization performance in tasks like image classification, it is not straightforward to extend existing data augmentation methods to the types of data used in robotic manipulation. Furthermore, most existing augmentation methods fall into one of two categories, and both have severe limitations:

In the first category, augmentations are defined by a set of transformations, sampled independently for each example. Most image augmentation methods fall into this category, where rotations or crops are sampled randomly for each example \cite{Augerino2020,AutoAugment,BestPractice2003}. By making augmentations independent of the example being augmented, we are restricted to operations which are valid on all examples. In the second category, there are methods which learn a generative model (VAE, GAN, etc.) of the data, and then sample new training examples from that model \cite{BayesianDATran2017,MaterialsAEOhno2020,PriceForecastingAE2021}. This approach assumes a useful generative model can be learned from the given dataset, but we found these methods did not perform well when the dataset is small.

\begin{figure}
    \centering
    \includegraphics[width=1\linewidth]{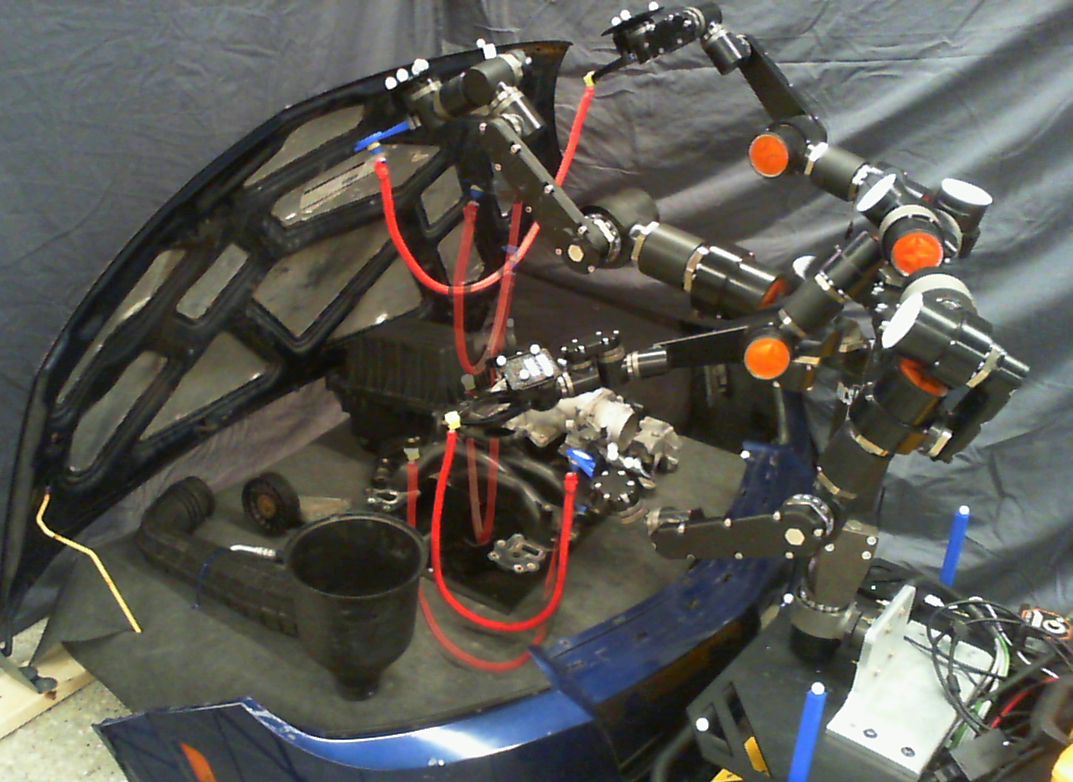}
    \caption{A mock-up of a car engine bay. The robot must move the rope and place it under the engine without snagging it to set up for lifting the engine. We use data augmentation to improve task success rate during online learning for this task.}
    \label{fig:real_robot_setup}
\end{figure}

\begin{figure*}
    \centering
    \includegraphics[width=1\linewidth]{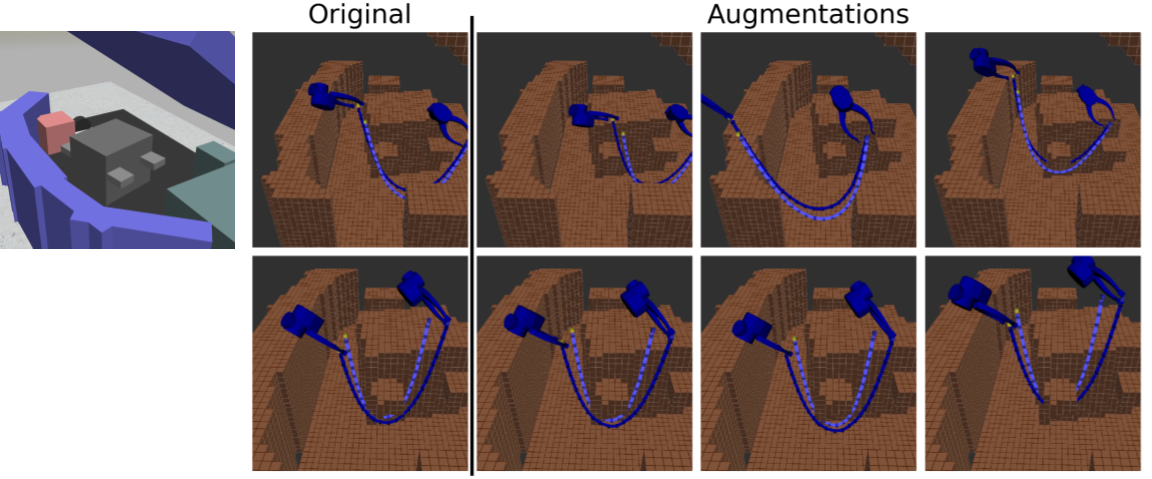}
    \caption{Examples of augmentations of rope generated by our method. On the left is a picture of the scene in simulation from a zoomed out viewpoint. The simplified engine block model is in the center. The rope start (dark blue) and end (light blue) states are shown, with the grippers shown at the start state. The static environment geometry is shown in brown. The first row shows a transition in free space, where the resulting augmentations are particularly diverse. The final augmentation shows how our method found a transformation to move the rope underneath the hook while remaining in free space. The second row shows a transition which involves contact between the rope and the environment. The augmentations preserve this contact.}
    \label{fig:rope_aug_examples}
\end{figure*}

It is not trivial to define a coherent framework for data augmentation that encompasses many domains and many types of learning problems (e.g. classification and regression). Thus, the first contribution of this paper is a formalization of the data augmentation problem. In our problem statement (Section \ref{sec:problem}), we formalize data augmentation as an optimization problem based on three key criteria: \textit{validity}, \textit{relevance}, and \textit{diversity}. We define an augmented example as \textit{valid} if it obeys the laws of physics and is possible in the real world. Augmentations are \textit{relevant} if they are similar to data that would be seen when performing the target task. \textit{Diversity} encourages the augmentations to be as varied as possible, i.e. the transformations applied to the data should be uniformly distributed to maximize diversity. Producing diverse augmentations for each original example is key to improving the generalization of the trained network.

The general definitions of validity, relevance, and diversity we propose depend on information that is intractable to compute for many manipulation problems, and therefore we also present approximations to these definitions. We do not claim that this formulation is useful for all manipulation problems, and clearly define the physical assumptions behind this formulation in Section \ref{sec:problem}.

Our second contribution is a method for solving this approximated optimization problem. Our method operates on trajectories of object poses and velocities, and searches for rigid-body transformations to apply to the moving objects in the scene to produce augmentations. Our method encourages validity by preserving contacts and the influence of gravity. Additionally, we encourage relevance by initializing the augmentations nearby the original examples and preserving near-contacts. Finally, we encourage diversity by pushing the augmentations towards randomly sampled targets.

Our results demonstrate that training on our augmentations improves downstream task performance for a simulated cluttered planar-pushing task and a simulated bimanual rope manipulation task. The learning problems in these tasks include classification and regression, and have high-dimensional inputs and outputs. Lastly, we demonstrate our augmentation in an online learning scenario on real-robot bimanual rope manipulation using noisy real-world perception data (Figure \ref{fig:real_robot_setup}). In this scenario, augmentation increased the success rate from 27\% to 50\% in only 30 trials. Additional materials such as code and video can be found on our \href{https://sites.google.com/view/data-augmentation4manipulation}{project website} \cite{ProjectWebsite}.

\section{Related Work}

This paper studies the problem of learning better models from limited data, which is important for robotics applications and has received significant attention as researchers have tried to apply deep learning to robotics \cite{DisbandOpenAI2021,ReviewKroemer2021}.

Data augmentation has been applied to many machine learning problems, from material science \cite{MaterialsAEOhno2020} to financial modeling \cite{PriceForecastingAE2021} (see \cite{TimeSeriesSurveyIwana2020,NLPSurveyFeng2021,ImageAugSurvey2019} for several surveys). It is especially common in computer vision \cite{ImageAugSurvey2019,BestPractice2003,AutoAugment,RLAugLaskin2020,Augerino2020}, and is also popular in natural language processing \cite{NLPSurveyFeng2021,NLPMa2019}. In these fields the data is often in standardized data types---images, text, or vectors of non-physical features (e.g. prices). Each of the data types can be used for a wide variety of tasks, and various data augmentations have been developed for various pairings of data type and task.

However, problems in robotic manipulation use other formats of data, such as point clouds, or object poses, and may consist of time-series data mixed with time-invariant data. To the best of our knowledge, there are currently no augmentation methods designed specifically for data of the types above. Our proposed method is intended to fill this gap.

In contrast to engineering augmentations based on prior knowledge, another body of work uses unsupervised generative models to generate augmentations \cite{BayesianDATran2017,MaterialsAEOhno2020,PriceForecastingAE2021}. Typically, these methods train a model like an Auto-Encoder or Generative Adversarial Network (GAN) \cite{GANGoodfellow14}) on the data, encode the input data into the latent space, perturb the data in the latent space, then decode to produce the augmented examples. These methods can be applied to any data type, and handle both regression and classification problems. However, they do not incorporate prior knowledge, and only add small but sophisticated noise. In contrast, we embed prior knowledge about the physical and spatial nature of manipulation, and as a result can produce large and meaningful augmentations, at the cost of being less generally-applicable.

\begin{figure*}
    \centering
    \includegraphics[width=1\linewidth]{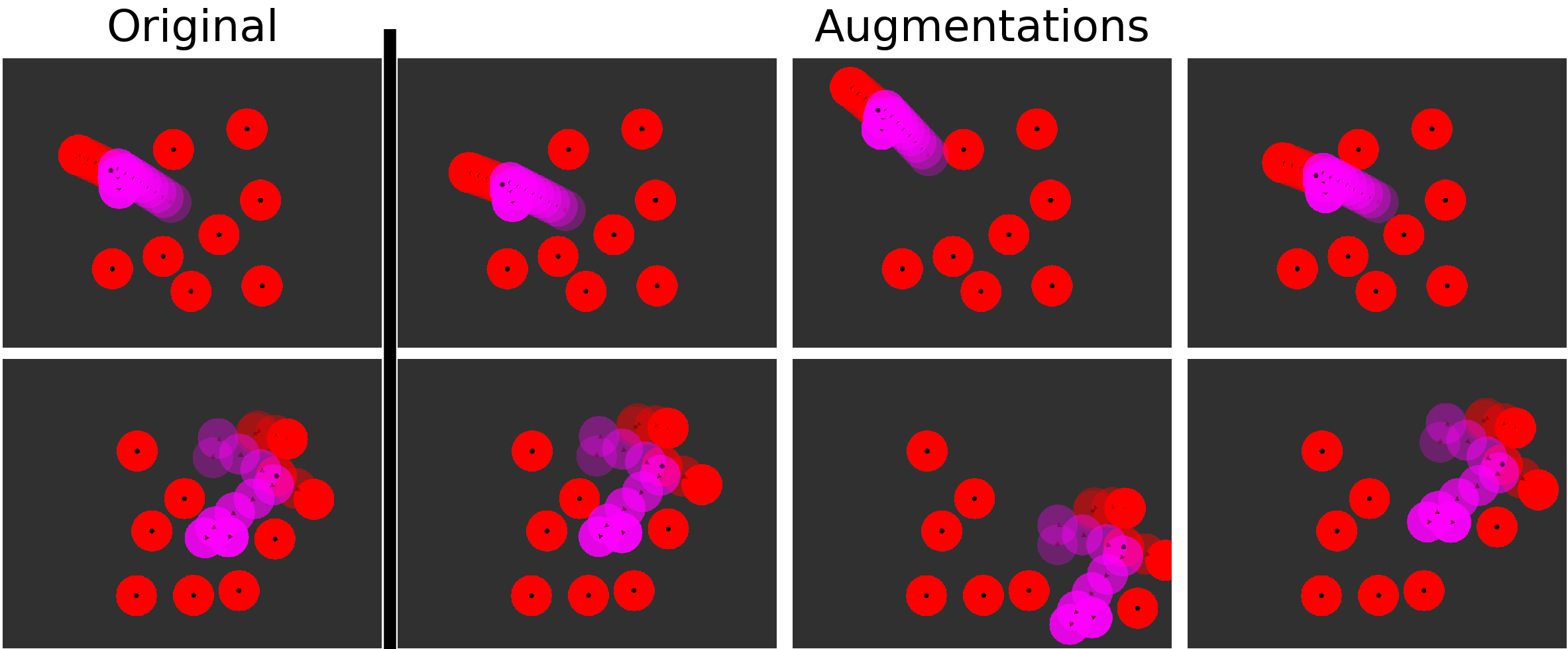}
    \caption{Examples of augmentations generated for learning the dynamics of planar pushing of 9 cylinders. The pink cylinder is the robot. Time is indicated by transparency. Augmentation transforms the positions and velocities of the cylinders that moved, including the robot. All moved objects are transformed together, rigidly. Despite the clutter, we are able to find relatively large transformations that still preserve existing contacts but do not create any new ones.}
    \label{fig:cylinders_aug_examples}
\end{figure*}

Although data augmentation is a popular approach, there are many other methods that have been proposed for data-efficient learning, both in general and specifically for robotic manipulation. We highlight a few important ones here, but note that these are all complementary to data augmentation. The most common technique is simply to pick a low-capacity model class, such as linear models or very small neural networks \cite{LinearAlarcon2013,TAMPC2021}. Alternatively, prior work has also developed various sets of priors specific to robotics \cite{RoboticPriors2015} which can be used as objectives during training. Another extremely useful technique is to engineer the state or action representations to include certain known invariances. For instance, a standard technique in dynamics learning methods is to represent the input positions in a local frame as opposed to a world frame, to encode translation invariance \cite{Propnet}. There are also methods for learning these kinds of invariances \cite{TAMPC2021}.

Finally, there are methods which automatically tune the parameters of a space of manually-defined augmentations, such as rotations and crops \cite{Augerino2020,AutoAugment}. These techniques are also compatible with our proposed method, and could be used to tune the hyperparameters of our algorithm.

\section{Problem Statement}
\label{sec:problem}

In this section, we formally define the form of data augmentation studied in this paper. We define a dataset $\dataset$ as a list of examples $\example\in\exampleSpace$ and, optionally, labels $\ell(\example)=\classLabel$, where $\ell$ is a task-specific labeling function. We assume the space $\exampleSpace$ is a metric space with a distance function $\distfname$. Augmentation is a stochastic function $\augDef$ which takes in an example $\example$ and produces the augmented example $\exampleAug$. The general form is shown in Algorithm \ref{alg:aug_prototype}. Internally, augmentation will call \sample{} to generate a vector of parameters, which we call $\transform$. We also define $\exampleAugSet$ as a set of $k$ augmented examples produced by calling \sample{} then \apply{} $k$ times. The parameters $\transform$ describe the transformation which will be applied to the example in the \apply{} procedure. We focus on augmentation functions that are stochastic, thus $\augf$ will sample new augmented examples each time it is called. If the dataset contains labels $\classLabel$, we assume that the labels should not change when the example is augmented.

\begin{algorithm}[t]
\caption{$\augf(x)$}\label{alg:aug_prototype}
$\transform = \sample{}(\example) $\\
$\aug{\example} = \apply(\example,\transform)$\\
return $\aug{\example}$\\
\end{algorithm}

We propose that useful augmentations should be \textit{valid}, \textit{relevant}, and \textit{diverse}. Let the valid set $\validSet$ be the set of examples which are physically possible. Let the relevant set $\relevantSet$ be the set of examples likely to occur when collecting data for or executing a specific set of tasks in a specific domain. We define $\mathrm{validity}(\exampleAug)=1$ if $\exampleAug \in \validSet$ and $\mathrm{validity}(\exampleAug)=0$ otherwise. We also define $\mathrm{relevance}(\exampleAug):=e^{-\distf{\exampleAug}{\relevantSet}}$ and $\mathrm{diversity}(\exampleAugSet):=\uniformity$, where $\klF$ is the Kullback–Leibler divergence and $\transformDist$ is the distribution of the parameters for a set of augmented examples $\exampleAugSet$. Diversity is maximized when the augmentation transformations are uniformly distributed in the range $\transformUniformRange$. With these concepts defined, we define data augmentation as the following optimization problem, the solution to which is a set of augmentations $\exampleAugSet$:

\begin{equation}
    \label{eq:most_general}
    \begin{array}{cc}
        \underset{\exampleAugSet}{\mathrm{max}} & \mathrm{diversity}(\exampleAugSet) + \beta \sum_{\exampleAug_i}\mathrm{relevance}(\exampleAug_i) \\
        \mathrm{subject\: to} & \mathrm{validity}(\exampleAug_i)\:\:\quad \forall\,\exampleAug_i \in \exampleAugSet\\
        & \ell(\aug{\example})=\ell(\example) \\
    \end{array}
\end{equation}

\noindent where $\beta$ is a positive scalar.

This optimization problem can be solved directly if $\validSet$, $\relevantSet$, and $\ell$ are known. However, in manipulation tasks, that is rarely the case. Instead, we will formulate an approximation to this problem using measures of relevance, diversity, and validity that are derived from physics and useful for a variety of robotic manipulation tasks and domains.

\begin{figure}
    \centering
    \includegraphics[height=3.85cm]{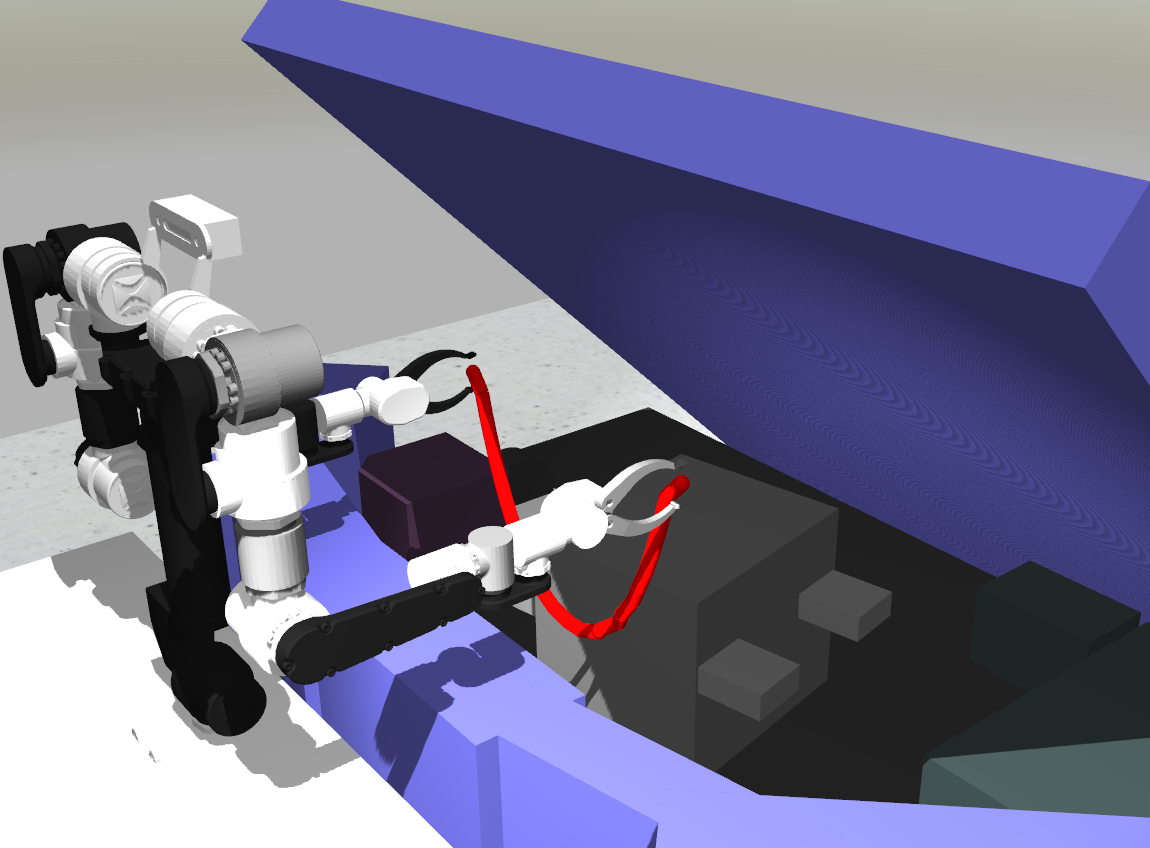}
    \includegraphics[height=3.85cm]{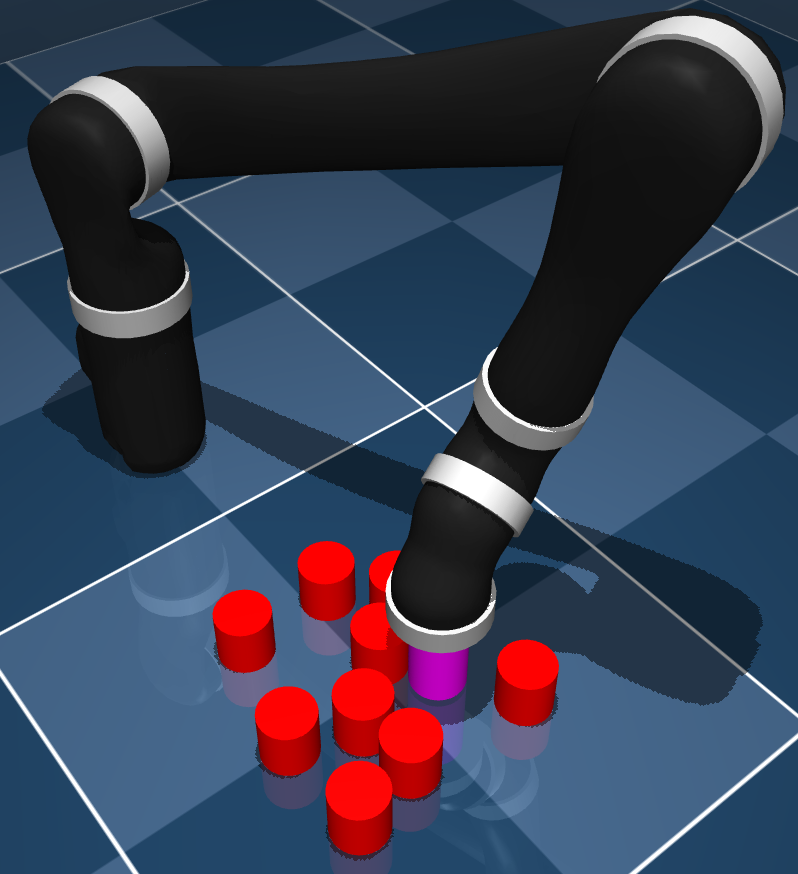}
    \caption{(left) The environment for bimanual rope manipulation, in simulation. (right) The environment for cluttered planar pushing of cylinders, in simulation.}
    \label{fig:sim_envs}
\end{figure}

\subsection{Assumptions}
\label{sec:assumptions}

Most augmentation algorithms rely on some expert knowledge or heuristics to define what is a valid augmentation. For instance, rotating an image for image classification makes an assumption that rotation does not change the label, and this is not always true. Similarly, the efficacy or correctness of our algorithm is also subject to certain assumptions. Here, we define the key assumptions:

\begin{itemize}
    \item The geometry of the robot and all objects is known.
    \item The scene can be decomposed into objects which can be assigned or detected as either moving or stationary.
    \item Examples are time-series, consisting of at least two states.
    \item All possible contacts between stationary vs. moving objects have the same friction coefficient.
    \item Contacts between the robot and objects/environment (e.g. grasps) can be determined from the data.
    \item A rigid-body transformation of an object preserves internal forces arising from its material properties.
    \item Objects only move due to contact or under the force of gravity. We do not handle movement due to magnetism or wind, for example.
\end{itemize}

Notably, the assumption that a rigid-body transformation preserves internal, material forces is what allows us to handle cluttered scenes with many moving objects, as well as deformable or articulated objects. While it could be valuable to augment the deformation or relative motion of the objects, doing so in a way that is valid would be challenging. Instead, we transform them all rigidly (See examples in Figures \ref{fig:rope_aug_examples},\ref{fig:cylinders_aug_examples}).

The assumption of having a common friction coefficient between all moving versus stationary objects is in-line with much manipulation research. For example, work on planar pushing assumes friction is uniform across the plane \cite{PushSim2021,PushYu2016}. Note that we make no assumption on the coefficients of friction between two moving objects.

Naturally, there are scenarios where these assumptions do not hold and thus where our algorithm may not perform well. However, our experiments demonstrate significant improvement on two very different manipulation scenarios, and we expect these assumptions extend to other scenarios as well.

\section{Methods}

We first describe an approximation to the augmentation problem \eqref{eq:most_general}, which is specialized for manipulation. Next, we decompose this problem and describe each component in detail.

\subsection{Algorithm Overview}

Since robotic manipulation is interested specifically in moving objects, we focus on augmenting trajectories of poses and velocities of moving objects. A key insight is that objects in the scene can be categorized as either robots, moved objects, or stationary objects, and that these should be considered differently in augmentation. We denote the moved objects state as $\state$, the robot state as $\robot$, the robot action as $\action$, and the stationary objects as $\env$ (also called environment). Our method augments the moved object states, the robot state, and the actions, but not the stationary objects. We do not assume any specific representation for states or actions, and examples of possible representations include sets of points, joint angles, poses, or joint velocities. Since we operate on trajectories, we bold the state ($\bm{\state},\bm{\robot}$) and action ($\bm{\action}$) variables to indicate time-series (e.g $\state_{1:T}=\bm{\state}$). With this categorization, we can write $\example=\{\bm{\state},\bm{\robot},\bm{\action},\env\}$ and $\aug{\example}=\{\bm{\aug{\state}},\bm{\aug{\robot}},\bm{\aug{\action}},\env\}$.

We choose the parameters $\transform$ to be rigid body transformations, i.e. either $SE(2)$ or $SE(3)$. We parameterize $\transform$ as a vector with translation and rotation components, with the rotation component with Euler angles bounded from $-\pi/2$ to $\pi/2$, which gives uniqueness and a valid distance metric. These rigid body transforms are applied to moved objects in the scene, and augmented robot state and action are computed to match. We choose rigid body transforms because we can reasonably assume that even for articulated or deformable objects, augmenting with rigid body transforms preserve the internal forces, and therefore the augmentations are likely to be valid.

It may seem that an effective method to generate augmentations is then to randomly sample transforms independent of the data. However, this is not an effective strategy because it is highly unlikely to randomly sample valid and relevant transformations. We confirm this in our ablations studies (included in the Appendix 1.A). Instead of sampling transforms randomly, we formulate an approximation to Problem \ref{eq:most_general}:

\begin{equation}
    \label{eq:method}
    \begin{array}{cc}
        \underset{\transform}{\mathrm{min}} & 
        \lossDiversity(\transform,\target{\transform}) +
        \betaBbox\lossBbox(\bm{\aug{\state}}) + \betaValid\lossValid(\transform) + \\
        & \betaOcc\lossOcc(\bm{\aug{\state}},e) + \betaDmd\lossDmd(\bm{\aug{\state}},\env) + \\
        & \lossRobot(\bm{\aug{\state}},\bm{\aug{\robot}},\bm{\aug{\action}},\env) \\[1em]
        \text{subject to} & \{\bm{\aug{\state}},\bm{\aug{\robot}},\bm{\aug{\action}},\env\} = \apply(\bm{\state},\bm{\robot},\bm{\action},\env,\transform) \\
        & \target{\transform}\sim\transformUniform \\
    \end{array}
\end{equation}

The decision variable is now the parameters $\transform$, and the validity constraint is moved into the objective. We propose that diversity should be maximized by the transforms being uniformly distributed, and therefore $\lossDiversity$ penalizes the distance to a target transform $\target{\transform}$ sampled uniformly within $[\transform^-,\transform^+]$. The relevance and validity terms (which are intractable to compute) are replaced with four objective functions, which are specialized to manipulation. The magnitudes of different terms are balanced by ${\betaBbox,\betaValid,\betaOcc,\betaDmd}$, which are defined manually. We define each objective function below:

\subsubsection{Bounding Box Objective}
First is the bounding-box objective $\lossBbox$, which keeps the augmented states $\bm{\aug{\state}}$ within the workspace/scene bounds defined by $[\state^-,\state^+]$. The bounding box objective encourages relevance, since states outside the workspace are unlikely to be relevant for the task.

\begin{equation}
\lossBbox = \sum_{i=1}^{|s|}{\max(0,\bm{\aug{\state}}_i - \state^+_i) + \max(0,\state^-_i-\bm{\aug{\state}}_i)}
\end{equation}

\subsubsection{Transformation Validity Objective}
The transformation validity objective $\lossValid$ assigns high cost to transformations that are always invalid or irrelevant for the particular task or domain. It is defined by function $\validF$, which takes in only the transformation. For example, in our rope manipulation case, it is nearly always invalid to rotate the rope so that it floats sideways. In our cluttered pushing task, in contrast, this term has no effect. This term can be chosen manually on a per-task basis, but we also describe how a transformation validity objective can be learned from data in section \ref{sec:learnValid}.

\begin{equation}
\lossValid = \validF(\transform)
\end{equation}

\subsubsection{Occupancy Objective}
The occupancy objective $\lossOcc$ is designed to ensure validity by preventing objects that were separate in the original example from penetrating each other and ensuring that any existing penetrations are preserved. In other words, we ensure that the occupancy $O(p)$ of each point $\augStatePoint\in\augStatePoints$ in the augmented object state matches the occupancy of the corresponding original point $\statePoint\in\statePoints$. For this term, we directly define the gradient, which moves $\augStatePoint$ in the correct direction when the occupancies do not match. This involves converting the environment $\env$ into a signed-distance field (SDF) and the moved objects states $\bm{\state}$ into points $\statePoints$. This objective assumes that the environment has uniform friction, so that a contact/penetration in one region of the environment can be moved to another region.

\begin{equation}
\lossOcc=\sum_{\substack{\statePoint\in\statePoints \\ \augStatePoint\in\augStatePoints}} \SDF(\augStatePoint)\big(O(\statePoint)-O(\augStatePoint)\big)
\end{equation}

\subsubsection{Delta Minimum Distance Objective}
The delta minimum distance objective is designed to increase relevance by preserving near-contact events in the data. We preserve near-contact events because they may signify important parts of the task, such as being near a goal object or avoiding an obstacle. We define the point among the moved object points $\statePoints$ which has the minimum distance to the environment $\pointMinDist=\mathrm{argmin}_{\statePoint}\SDF(\statePoint)$. The corresponding point in the augmented example we call $\augPointMinDist$.

\begin{equation}
\lossDmd = ||\SDF(\pointMinDist)-\SDF(\augPointMinDist)||^2_2
\end{equation}

\begin{figure}
    \centering
    \includegraphics[width=1\linewidth]{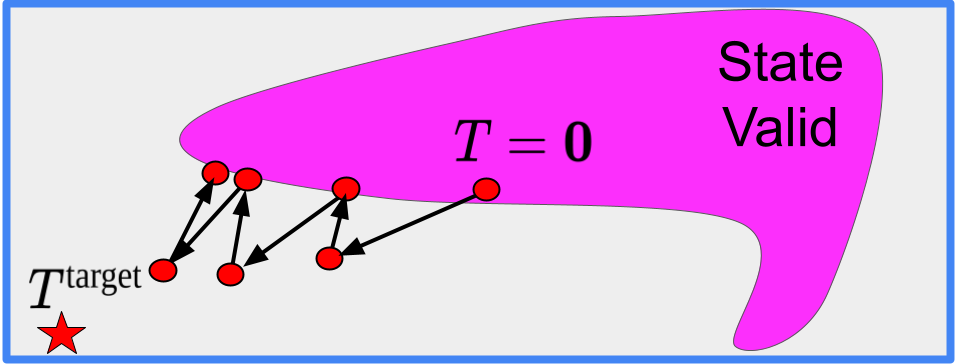}
    \caption{Illustration of \augState{} within Algorithm \ref{alg:aug}. All points and sets are in the space of $\transform$. The path of $\transform$ is shown in red with black arrows. The pink set, State Valid, is the set where \textit{state\_valid} is true. $\transform$ begins at the origin, and alternates between moving towards $\target{\transform}$ and projecting back into the set \textit{state\_valid} (by solving Equation \eqref{eq:project}).}
    \label{fig:step_project}
\end{figure}

\subsubsection{Robot Contact Objective}

The robot contact objective $\lossRobot$ ensures validity of the robot's state and the action. This means that contacts involving the moved objects which existed in the original example must also exist in the augmented example. Let the contact points on the robot be $\contactpoint_{\robot}$ and the contact points on the moved objects' state be $\contactpoint_{\state}$.

\begin{equation}
\lossRobot=\sum_i(||{\contactpoint_{\robot,i} - \contactpoint_{\state,i}}||^2_2)
\end{equation}

Finally, we note that other objective functions can be added for the purpose of preserving task-specific labels, i.e. so that $\ell(\example)=\ell(\aug{\example})$. However, for our experiments, no additional functions were necessary.

\subsection{Solving the Augmentation Optimization Problem}

This section describes how we solve Problem \eqref{eq:method}, and the procedure is detailed in Algorithm \ref{alg:aug}. First, we split the problem into two parts, \augState{} and \augRobot{}.

\begin{algorithm}[t]
\caption{$\augf(\bm{\state},\bm{\robot},\bm{\action},\env)$}\label{alg:aug}
\SetKwComment{Comment}{// }{}
\Comment{\augState{}}
$ \target{\transform} \sim \transformUniform $\\
$\transform = \mathbf{0}$ \text{ (identity)} \\
\For{$i \in \nStep$}{
    $\transform_\text{old}=\transform$ \\
    $\transform = \texttt{step\_towards}(\transform, \target{\transform})$ \\
    $\transform = $ solve Equation \eqref{eq:project} \\
    \If{$\distf{\transform} {\transform_\text{old}}<\projectionNotProgressing$}{
        break\\
    }
}
\Comment{\augRobot{}}
$\bm{\aug{\state}}, \textit{state\_valid} = \applyState(\bm{\state},\bm{\action},\transform)$\\
$\bm{\aug{\robot}},\bm{\aug{\action}}$, \textit{ik\_valid} $\leftarrow \mathrm{IK}(\bm{\robot},\bm{a},\bm{\aug{\state}},\env)$\\
\uIf{$!$state\_valid or $!$ik\_valid}{
    return $\bm{\state},\bm{\robot,}\bm{\action},\env$\\
}
\Else {
    return $\bm{\aug{\state}},\bm{\aug{\robot}},\bm{\aug{\action}},\env$\\
}
\end{algorithm}

In \augState{}, we optimize the transform $\transform$ to produce the moved objects' state $\bm{\aug{\state}}$ while considering environment $\env$. To achieve diversity, we uniformly sample a target transform $\target{\transform}$ and step towards it iteratively. This stepping alternates with optimizing for validity and relevance. We visualize this procedure in Figure \ref{fig:step_project}, as well as in the supplementary video. The innermost optimization problem is 
\begin{equation}
    \label{eq:project}
    \begin{array}{cc}
        \underset{\transform}{\mathrm{argmax}} & \betaBbox\lossBbox+\betaValid\lossValid+\betaOcc\lossOcc+\betaDmd\lossDmd \\
    \end{array}
\end{equation}

We solve Problem \eqref{eq:project} using gradient descent, terminating after either $\mProj$ steps or until the gradient is smaller than some threshold $\projStepSizeThreshold$.

Note that we start \augState{} in Algorithm \ref{alg:aug} with $\transform$ at the identity transformation, rather than initially sampling uniformly. This has two benefits. First, the identity transform gives the original example, which is always in the relevant set. Second, it is unlikely that a uniformly sampled transformation is valid or relevant, so starting at a random transformation would make solving Problem \eqref{eq:project} more difficult.

In \augRobot{}, we are optimizing $\lossRobot$. This corresponds to computing the augmented robot states $\bm{\aug{\robot}}$ and actions $\bm{\aug{\action}}$ given the augmented states $\bm{\aug{\state}}$ and the environment $\env$. Minimizing $\lossRobot$ means preserving the contacts the robot makes with the scene, which we do with inverse kinematics (Line 10 in Algorithm \ref{alg:aug}).

\subsection{Learning the Valid Transforms Objective}
\label{sec:learnValid}

As discussed above, we include a term $\lossValid$ based only on the transformation $\transform$. In some cases, such as our rope manipulation example, it may not be obvious how to define this objective manually. Our rope is very flexible, and therefore rotating the rope so that it floats in a sideways arc is invalid, but it may be valid for a stiff rope or cable. To address this, we offer a simple and data efficient algorithm for learning the transformation validity function $\validF$.

Our method for learning $\validF$ is given in Algorithm \ref{alg:valid_transformations_data}. This algorithm repeatedly samples augmentations of increasing magnitude, and tests them on the system (lines 6 and 8). This generates ground truth states starting from an input state and action. The result is a dataset $\learnValidDataset$ of examples ($\transform$, $\learnValidError$). We then train a small neural network $\learnedValidF$ to predict the error $\learnValidError$ and use the trained model as our transformation validity objective. We collect $\nLearnValid = \sqrt{10^\transformDim}$ examples, where $\transformDim$ is the dimensionality of the space of the transformation $\transform$.

\begin{algorithm}[t]
\caption{Data Collection for Learning Valid Transformations}\label{alg:valid_transformations_data}
\SetKwComment{Comment}{/* }{ */}
\KwIn{$\learnValidStateActionSet,\nLearnValid$}
\KwOut{$\learnValidDataset$}
$\minLearnValidError = \infty$ \\
\For{$i \in [1, \nLearnValid]$}{
    \For{$(\state_t,\robot_t,\action_t,\env) \in \learnValidStateActionSet$}{
        $ \learnValidScaling = i / \nLearnValid $ \\
        $ \transform \sim \mathbb{U}[\learnValidScaling\transform^-,\learnValidScaling\transform^+] $\\
        $ \state_{t+1}, \robot_{t+1} = \learnValidSimF(\state_t, \robot_t, \action_t, \env)$ \\
        $ \aug{\state}_{t,t+1}, \aug{\robot}_{t,t+1}, \aug{\action}_{t} = \apply(\state_{t,t+1},\robot_{t,t+1},\action_{t},\transform)$ \\
        $ \test{\aug{\state}_{t+1}}, \test{\aug{\robot}_{t+1}} = \learnValidSimF(\aug{\state_t}, \aug{\robot_t}, \aug{\action}_t, \env)$ \\
        $\learnValidError = || \aug{\state}_{t+1} - \test{\aug{\state}_{t+1}} ||$ \\
        \If {$\learnValidError < \minLearnValidError$}{
            $\minLearnValidError = \learnValidError,\,\transform_\text{min} = T,\,\minLearnValidError = \learnValidError$ \\
        }
    }
    add $(\transform_\text{min}, \minLearnValidError)$ to $\learnValidDataset$ \\
}
return $\learnValidDataset$
\end{algorithm}

This method owes its efficiency and simplicity to a few key assumptions about the system/data. First, we assume that we can collect a few ($<1000$) examples from the system and test various transformations. This could be performed in a simulator, as we do in our experiments. Because the transformation validity objective is not a function of state, action, or environment, we can make simplifications to this simulation by picking states and environments which are easy to simulate. We denote this set of states and actions as $\learnValidStateActionSet$. Second, because the transformation parameters are low-dimensional (3 and 6 in our experiments) the trained model generalizes well with relatively few examples.

\subsection{Application to Cluttered Planar Pushing}

In this section, we describe how we apply the proposed method to learning the dynamics of pushing of 9 cylinders on a table (Figure \ref{fig:sim_envs}). The moved object state $\state$ consists of the 2D positions and velocities of the cylinders. The robot state $\robot$ is a list of joint positions, and the actions $\action$ are desired end effector positions in 2D. There is no $\classLabel$ in this problem. The parameters $\transform$ used are $SE(2)$ transforms. In this problem, any individual trajectory may include some moved cylinders and some stationary ones. In our formulation, the stationary cylinders are part of $\env$ and are not augmented, whereas the moved ones are part of $\state$ and are augmented. The robot's end effector (also a cylinder) is also augmented, and IK can be used to solve for joint configurations which match the augmented cylinders' state and preserve the contacts between the robot and the moved cylinders.

\subsection{Application to Bimanual Rope Manipulation}

In this section, we describe how we apply the proposed method to a bimanual rope manipulation problem (Figure \ref{fig:sim_envs}). In this problem, there is a binary class label, so $\classLabel\in\{0,1\}$, which is preserved under our augmentation (last constraint in Problem \eqref{eq:most_general}). The rope is the moved object, and its state $\state$ is a set of 25 points in 3D. The robot state $\robot$ is a list of the 18 joint positions, and the actions $\action$ are desired end effector positions in the robot frame. In this problem, we know that the only contacts the robot makes with the objects or environment are its grasps on the rope. Therefore, we preserve these contacts by solving for a robot state and action that match the augmented points on the rope. The parameters $\transform$ used are $SE(3)$ transforms.

\section{Results}

\begin{figure}
    \centering
    \includegraphics[width=1\linewidth]{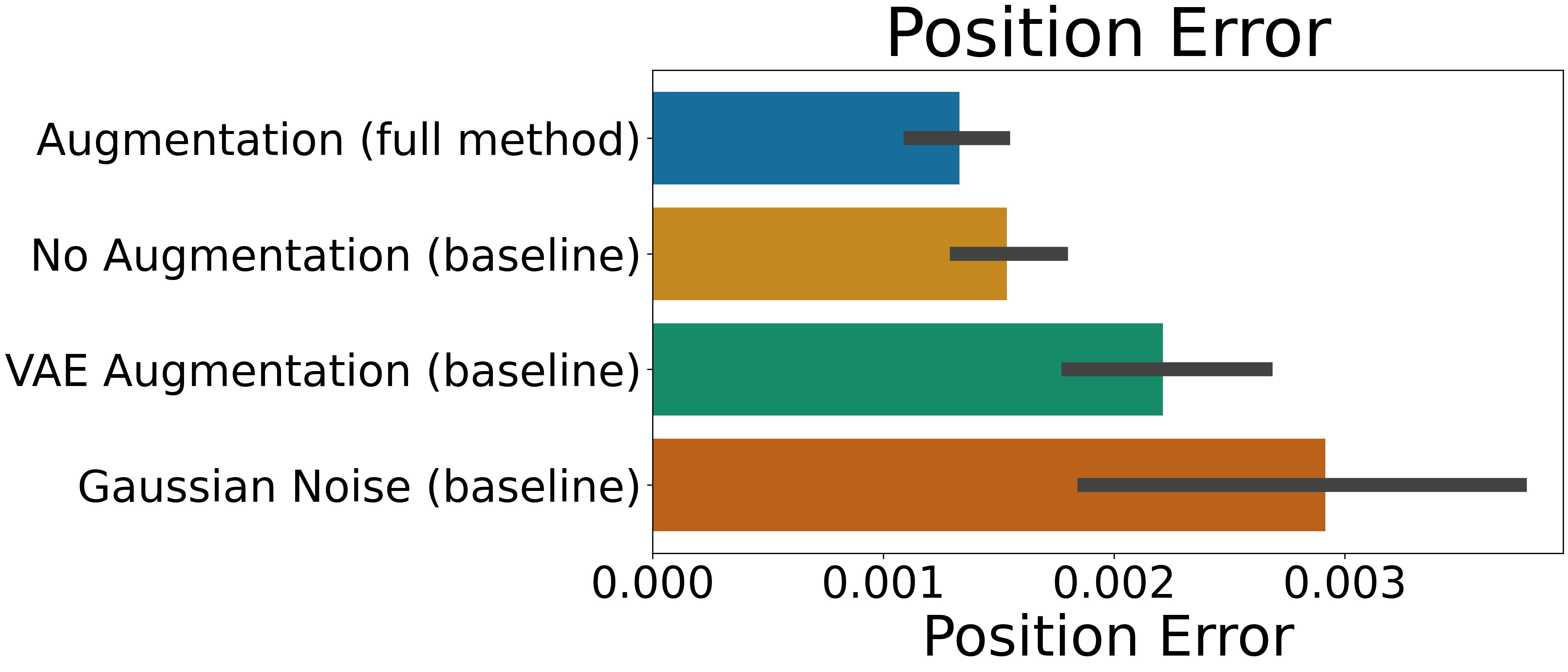}
    \caption{Mean position error (meters) for learning the dynamics of cluttered planar pushing.}
    \label{fig:cylinders_results}
\end{figure}

We start by describing the tasks and our experimental methodology, then we present our results. These experiments are designed to show that training on augmentations generated by our method improves performance on a downstream task. We perform two simulated experiments, where we run thousands of evaluations, including several ablations (see Appendix 1.A). We also perform a real robot experiment (Figure \ref{fig:real_robot_setup}) where we run 30 iterations of online validity classifier learning, with augmentation and without. In all experiments, we train until convergence or for a fixed number of training steps. This ensures a fair comparison to training without augmentation, despite the differing number of unique training examples. In all experiments we generate 25 augmentations per original example (See Appendix 1.B). We define key hyperparameters of our method in Appendix 1.C. A link to our code is available on the \href{https://sites.google.com/view/data-augmentation4manipulation}{project website} \cite{ProjectWebsite}.

\subsection{Cluttered Planar Pushing}

The cluttered planar pushing environment consists of a single robotic arm pushing 9 cylinders around on a table. The task is to learn the dynamics, so that the motion of the cylinders can be predicted given initial conditions and a sequence of robot actions. For this, we use PropNet \cite{Propnet}, and our task is inspired by the application of PropNet to planar pushing in \cite{DBRP2020}. The inputs to PropNet are an initial state $\state_0$ and a sequence of actions $\bm{\action}$, and the targets are the future state $(\state_1,\dots,\state_T$). All trajectories are of length 50. We evaluate the learned dynamics by computing the mean and maximum errors for position and velocity on a held-out test set. Example augmentations for this scenario are shown in Figure \ref{fig:cylinders_aug_examples}. 

This is an interesting application of our augmentation for several reasons. First, it is a regression task, which few augmentation methods allow. Second, the output of the dynamics network is high-dimensional (900 for a prediction of length 50), which normally means large datasets are needed and engineering invariances into the data or network is difficult. Finally, the trajectories contain non-negligible velocities and are not quasi-static.

\begin{figure}
    \centering
    \includegraphics[width=1\linewidth]{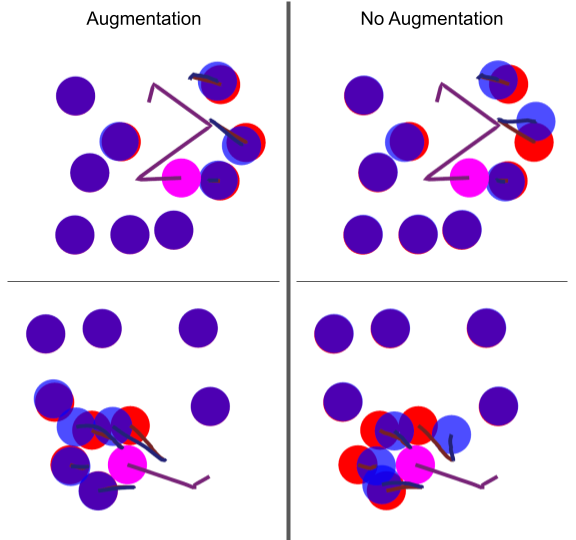}
    \caption{Predictions (blue) vs. ground truth (red) for planar pushing. The robot is in pink. Trajectories are visualized with lines. The left column shows predictions from a model trained with augmentation, the right column without.}
    \label{fig:cylinders_rollouts}
\end{figure}

The original dataset contained 60 trajectories of length 50, or \SI{3000} time steps in total. For comparison, previous work on the same dynamics learning problem used over \SI{100000} time steps in their datasets \cite{Propnet,DBRP2020}. This is similar to the number of training examples we have \textit{after} augmentation, which is \SI{75000} time steps. Finally, we measured the performance of our implementation and found that for the planar pushing scenario we generate 4.5 augmentations per second on average.

The primary results are shown in Figure \ref{fig:cylinders_results}. Augmentation reduces the average position error from \SI{0.00154}{\meter} to \SI{0.00133}{\meter}, a decrease of 14\%. Additionally, we include two baselines, one which adds Gaussian noise to the state, robot, action, and environment data, and one which uses a VAE to generate augmentations as in \cite{MaterialsAEOhno2020}. The magnitude of the Gaussian noise was chosen manually to be small but visually noticeable. Our proposed augmentation method is statistically significantly better than the baseline without augmentation ($p<0.0362$), the Gaussian noise baseline ($p<0.0001$), and the VAE baseline ($p<0.0002$). This difference in error may seem small, but note that error is averaged over objects, and most objects are stationary. Two roll-outs from with-augmentation and from without-augmentation are shown in Figure \ref{fig:cylinders_rollouts}. In particular, we found that augmentation reduces ``drift,'' where the model predicts small movements for objects that should be stationary. Finally, we note that the Gaussian noise and VAE baselines perform worse than no augmentation, suggesting that data augmentation can hurt performance if the augmentations are done poorly.

\subsection{Bimanual Rope Manipulation}

In this task, the end points of a rope are held by the robot in its grippers in a scene resembling the engine bay of a car, similar to \cite{UnreliableMitrano2021}, and shown in Figure \ref{fig:sim_envs}. The robot has two 7-dof arms attached to a 3-dof torso with parallel-jaw grippers. The tasks the robot performs in this scene mimic putting on or taking off lifting straps from the car engine, installing fluid hoses, or cable harnesses. These tasks require moving the strap/hose/cable through narrow passages and around protrusions to various specified goal positions without getting caught. One iteration consists of planning to the goal, executing open-loop, then repeating planning and execution until a timeout or the goal is reached. The goal is defined as a spherical region with \SI{4.5}{\centi\meter} radius, and is satisfied when any of the points on the rope are inside this region.

The planner is an RRT with a learned constraint checker for edge validity (validity classifier), and more details are given in \cite{UnreliableMitrano2021}. We want to learn a classifier that takes in a single transition $x = (\state_t,\action_t,\state_{t+1},\env_t)$ and predicts whether the transition is valid. Without a good constraint checker, the robot will plan trajectories that result in the rope being caught on obstacles or not reaching the goal. We apply our augmentation algorithm to the data for training this constraint checker. After an execution has completed, the newly-collected data along with all previously collected data are used to train the classifier until convergence. Example augmentations for this scenario are shown in Figure \ref{fig:rope_aug_examples}. The objective is to learn the constraint checker in as few iterations as possible, achieving a higher success rate with fewer data.

\begin{figure}
    \centering
    \includegraphics[width=1\linewidth]{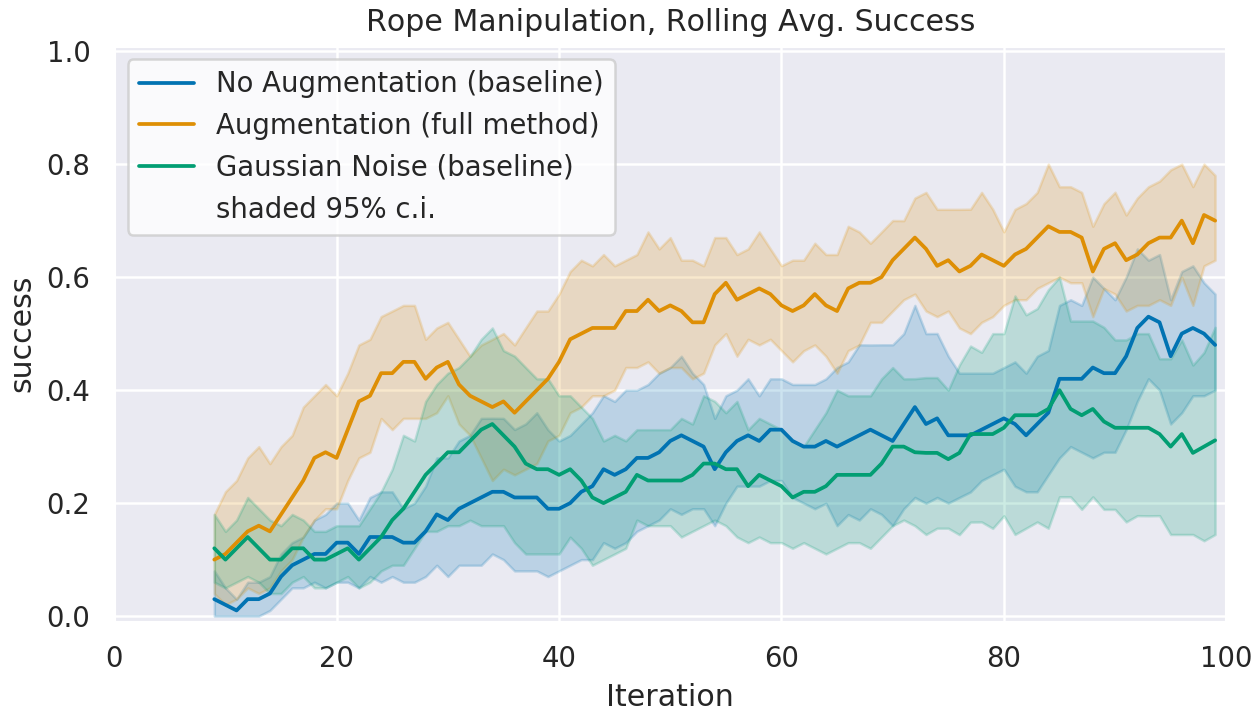}
    \caption{The success rate on simulated bimanual rope manipulation, using a moving window average of 10.}
    \label{fig:rope_results}
\end{figure}

In this experiment, a total of \SI{3038} examples were gathered (before augmentation, averaged over the 10 repetitions). Since the purpose of our augmentations is to improve performance using small datasets, it is important that this number is small. In contrast, prior work learning a similar classifier used over \SI{100000} examples in their datasets \cite{UnreliableMitrano2021,UnreliableDale2019}. This is similar to the number of training examples we have \textit{after} augmentation, which is \SI{75950} on average.  Finally, we measured the performance of our implementation and found that for the rope scenario we generate 27 augmentations per second on average.

The primary results are shown in Figure \ref{fig:rope_results}. Over the course of 100 iterations, the success of our method using augmentation is higher than the baseline of not using augmentation, as well as the Gaussian noise baseline. We omit the VAE baseline, since it performed poorly in the planer pushing experiment. Furthermore, it is computationally prohibitive to retrain the VAE at each iteration, and fine-tuning the VAE online tends to get stuck in bad local minima. The shaded regions show the 95th percentile over 10 runs. If we analyze the success rates averaged over the final 10 iterations, we find that without augmentation the success rate is 48\%, but with augmentation the success rate is 70\%. The Gaussian noise baseline has a final success rate of 31\%. A one-sided T-test confirms statistical significance ($p<0.001$ for both).

\subsection{Real Robot Results}
\label{sec:real_robot_results}

\begin{figure}
    \centering
    \includegraphics[width=1\linewidth]{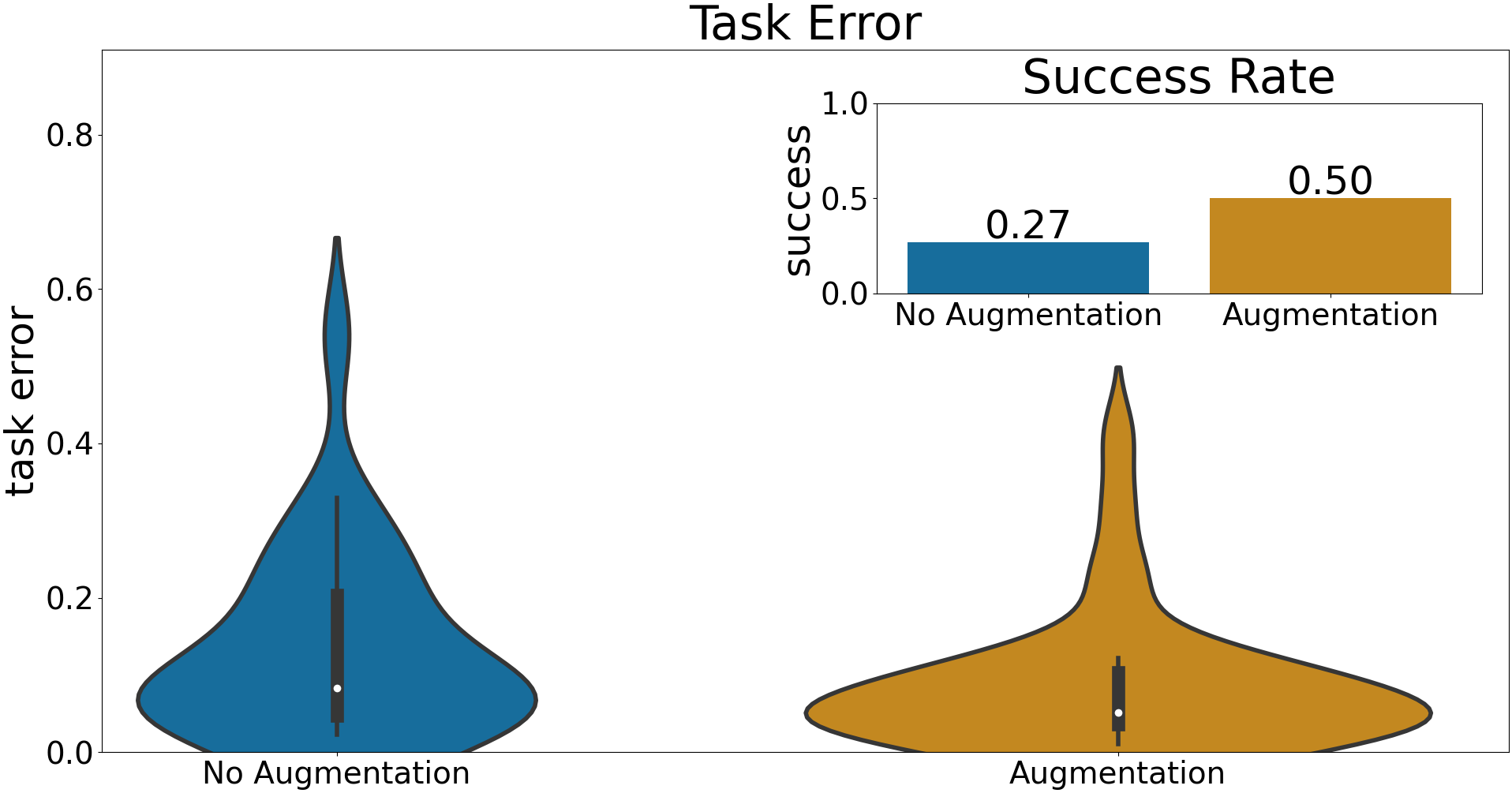}
    \caption{The success rate and task error distribution of bimanual rope manipulation on the real robot. Task error is the distance between the goal and the final observed state of the rope.}
    \label{fig:real_robot_results}
\end{figure}

In this section, we perform a similar experiment to the simulated bimanual rope manipulation experiment, but on real robot hardware. This demonstrates that our method is also effective on noisy sensor data. More importantly, it demonstrates how augmentation enables a robot to quickly learn a task in the real world. We use CDCPD2 \cite{CDCPD2} to track the rope state. The geometry of the car scene is approximated with primitive geometric shapes, like in the simulated car environment.

We ran the validity classifier learning procedure with a single start configuration and a single goal region, both with and without augmentation. After 30 iterations of learning, we stop and evaluate the learned classifiers several times. With augmentation, the robot successfully placed the rope under the engine 13/26 times. Without augmentation, it succeeded 7/26 times. The Gaussian noise and VAE baselines performs poorly in simulated experiments, therefore we omit them in the real robot experiments.

\section{Limitations}

There are problems and applications where the proposed objective functions do not ensure validity, relevance, and diversity. In these cases, the structure of our augmentation and projection procedures can remain, while new objective functions are developed. Another limitation is that our method is not compatible with image data. Much recent research in robotics has moved away from engineered state representations like poses with geometric information, and so there are many learning methods which operate directly on images. Although this is a limitation of the proposed method, many of the augmentations developed for images are also not applicable to problems in manipulation, even when images are used. For instance, pose detection, 3D reconstruction, semantic segmentation, and many other tasks may not be invariant to operations like cropping, flipping, or rotating. Creating an augmentation method for manipulation that is applicable to images is an open area for future research.

\section{Conclusion} 
\label{sec:conclusion}

This paper proposes a novel data augmentation method designed for trajectories of geometric state and action robot data. We introduce the idea that augmentations should be valid, relevant, and diverse, and use these to formalize data augmentation as an optimization problem. By leveraging optimization, our augmentations are not limited to simple operations like rotation and jitter. Instead, our method can find complex and precise transformations to the data that are valid, relevant, and diverse. Our results show that this method enables significantly better downstream task performance when training on small datasets. In simulated planar pushing, augmentation decreases the prediction error by 14\%. In simulated bimanual rope manipulation, the success rate with augmentation is 70\% compared to 47\% without augmentation. We also perform the bimanual rope manipulation task in the real world, which demonstrates the effectiveness of our algorithm despite noisy sensor data. In the real world experiment, the success rate improves from 27\% to 50\% with the addition of augmentation.

\section*{Acknowledgments}
The authors would like to acknowledge Andrea Sipos for her ingenious design of a reset mechanism, allowing us to run robot experiments unattended. This work was supported in part by NSF grants IIS-1750489 and IIS-2113401, ONR grant N00014-21-1-2118, and the Toyota Research Institute. This article solely reflects the opinions and conclusions of its authors and not TRI or any other Toyota entity.

\bibliographystyle{plainnat}
\bibliography{references}

\end{document}


\maketitle

\section{Appendix}
\subsection{Ablation of Objective Terms}
\label{sec:ablations}

\begin{figure}[h]
    \centering
    \includegraphics[width=1\linewidth]{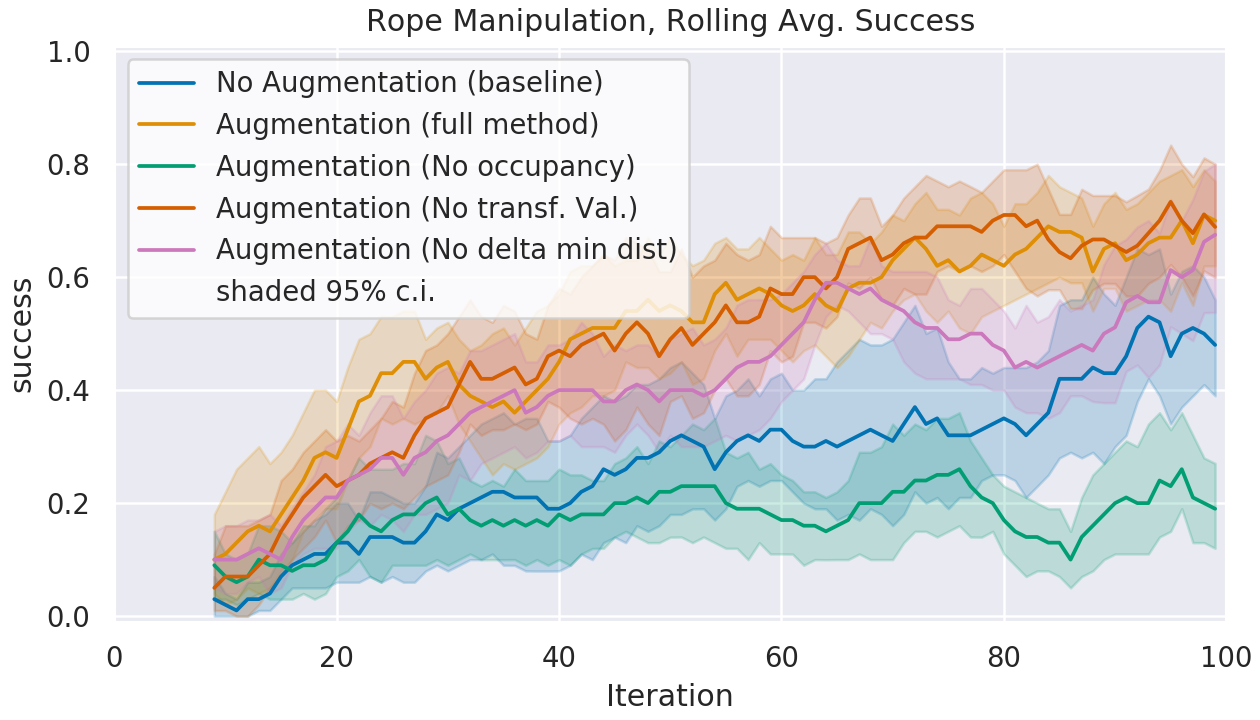}
    \caption{Success vs Iterations for ablations. Bimanual rope manipulation, in simulation.}
    \label{fig:rope_ablations}
\end{figure}

The objectives terms we propose are designed to be useful in many cases, but to understand this better, we ablate several of the objective terms. We evaluate the importance of the transformation validity objective, occupancy objective, and delta minimum distance objective by repeating our experiments. In each ablation, we omit one objective term. Each condition was run 10 times with different random seeds. The results are shown in Table \ref{tab:ablations} and Figure \ref{fig:rope_ablations}.

\begin{table}[h]
    \centering
    \begin{tabular}{l|l|l}
        Method & \thead{Task Success \% ($\uparrow$) \\ (rope)} & \thead{Position Error (m) ($\downarrow$) \\ (planar pushing)} \\
        Full Method & 0.700 (0.118) & 0.0010 (0.0001) \\
        No transf. valid. & 0.700 (0.185) & 0.0011 (0.0001) \\
        No delta min dist & 0.675 (0.185) & 0.0012 (0.0002) \\
        No occupancy & 0.24 (0.136) & 0.0028 (0.0011) \\
    \end{tabular}
    \caption{Ablations: Metrics on both tasks, with various objective terms removed. The standard deviation is shown in parentheses.}
    \label{tab:ablations}
\end{table}

We find that across both experiments, the most important objective was the occupancy objective. Without this objective, our method produced augmentations where contacts and penetrations are not preserved. This is invalid, and training on these examples produces a worse model even than using no augmentations.

\begin{figure}
    \centering
    \includegraphics[width=0.492\linewidth]{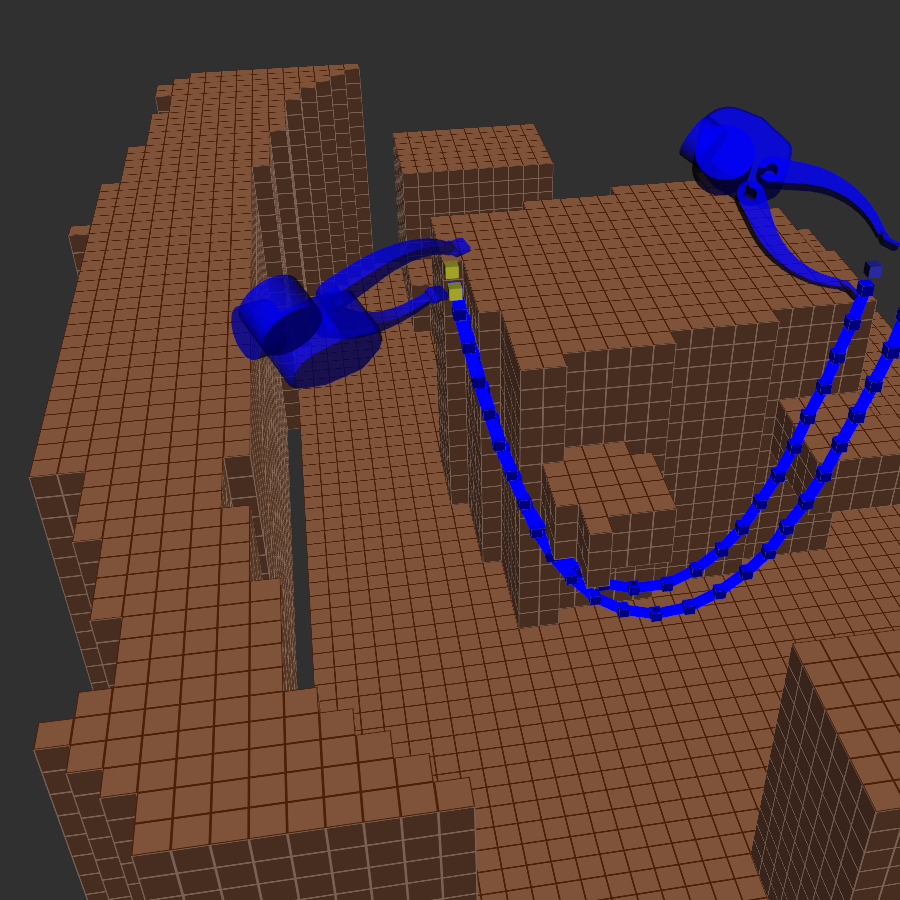}
    \includegraphics[width=0.492\linewidth]{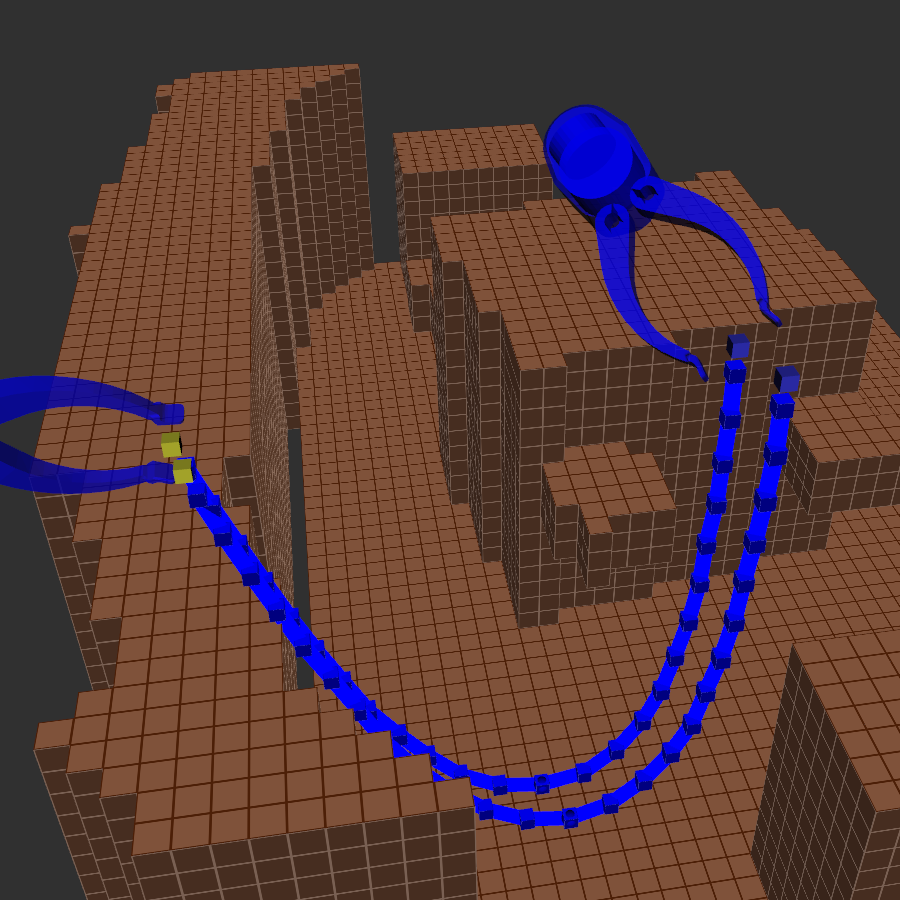}
    \caption{Augmented rope data generated without the delta minimum distance objective. On the left is the original transition, under the hook. On the right is the augmented transition, which is far from the hook. Without the delta minimum distance objective, data can be augmented away from interesting regions, becoming less relevant. This is based on the heuristic that contact and near-contact events are relevant for manipulation.}
    \label{fig:rope_dmd_comparison}
\end{figure}

The next most important objective is the delta minimum distance objective. In our simulated rope experiment, the method without this objective performs slightly worse. By visualizing the different examples (Figure \ref{fig:rope_dmd_comparison}), we found that without this objective, the augmentation transforms examples which were near the hooks and protrusions far into free space. In contrast, with the delta minimum distance objective, these examples stay near the hooks. Since the planning tasks involve moving in narrow passages around these hooks, it's more relevant to produce more examples in this area. Hence, by preserving the distance to nearby obstacles, we also tend to keep examples in areas of interest. In the planar pushing task, however, omitting the delta minimum distance objective did not have a notable effect. This could be in part due to the cluttered nature of the scene, which means that most transformations are small.

Finally, we find that the valid transformation objective can be omitted without effecting performance in these two tasks. In the planar pushing case this is completely expected, because there are no SE(2) transformations which are always invalid or irrelevant. However, in the rope experiment, this result is less intuitive. We found that omitting this objective occasionally produced augmentations where the rope is floating sideways in physically impossible states. This result suggests that even if a few augmentations are invalid or irrelevant, training on the augmentations still significantly outperforms using no augmentations. Even if omitting the objective does not degrade performance in these experiments, it is a natural term to include and may be beneficial in other applications.

\subsection{Choosing the Number of Augmentations}

\begin{figure}[ht]
    \centering
    \includegraphics[width=1\linewidth]{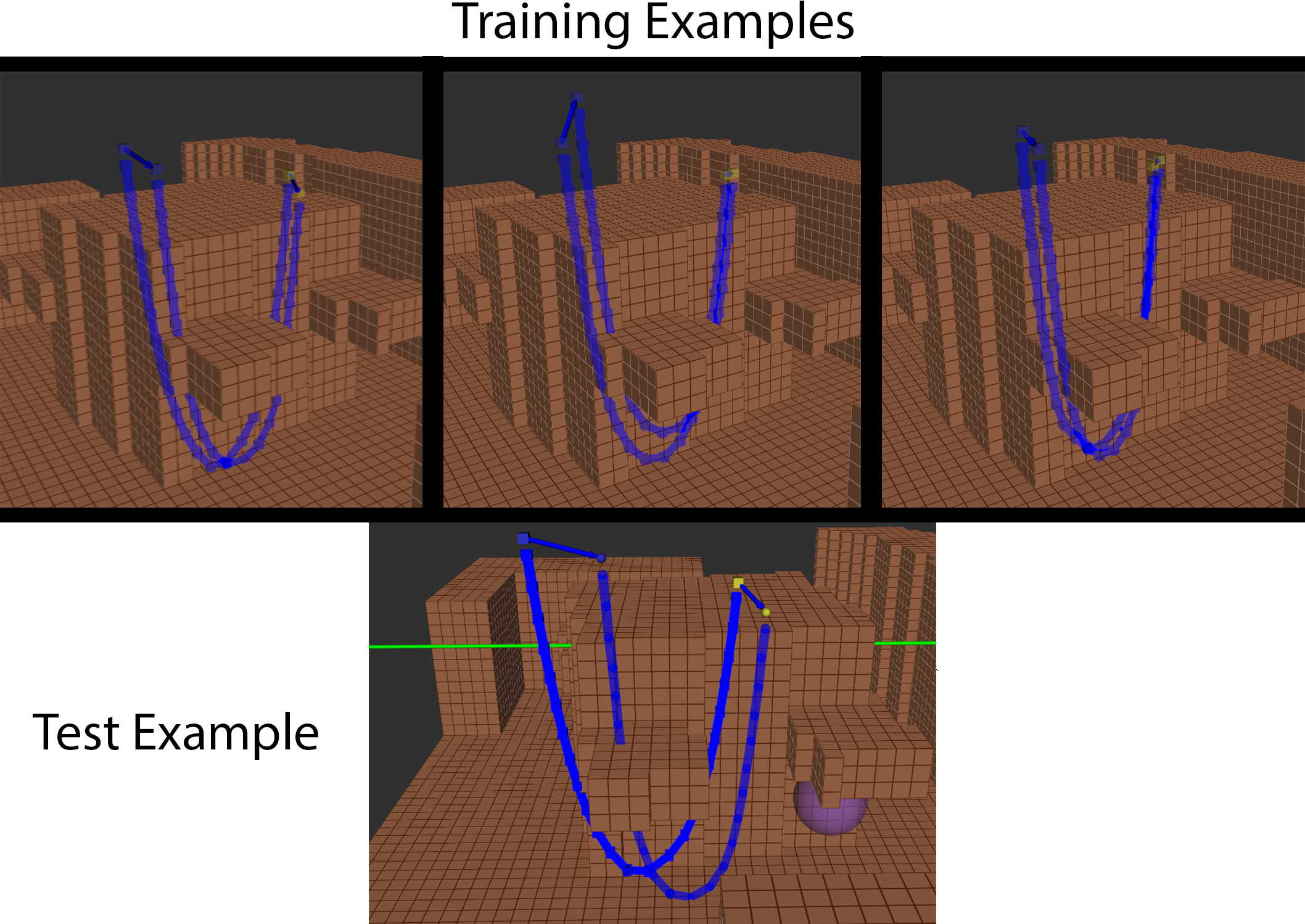}
    \caption{(top) These three original training example show rope transitions with inaccurate predictions, where the rope is predicted to move inside the hook on the engine. (bottom) The test transition is similar, but is not identical. The proposed augmentation method can move the rope transition while ensuring it still intersects the hook in a similar way, which allows it to generate augmentations like the test example.}
    \label{fig:num_augs_examples}
\end{figure}

Our method produces multiple distinct augmentations for each original example. More augmentations should improve generalization after training, but at the cost of additional computation time. In this experiment, we explore how the number of augmentations per-example affects generalization. We tested this in the rope domain on the task of learning the constraint checker. The original un-augmented dataset came from the first three iterations of learning. With these examples, we then generated varying numbers of augmentations for each example and evaluated on a held-out test example drawn from the fourth iteration. Figure \ref{fig:num_augs_examples} shows three rope transitions labeled 0 from the original three training examples, followed by the held-out test example. Intuitively we would expect, with enough augmentations, that from these original training examples we would be able to generate augmented examples which are very similar to the held-out test example.

\begin{figure}[ht]
    \centering
    \includegraphics[width=1\linewidth]{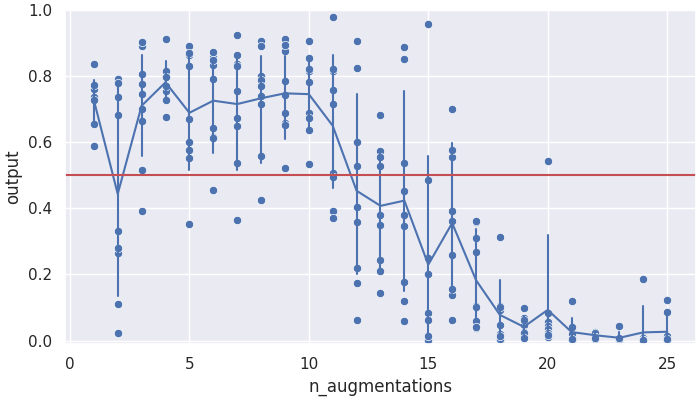}
    \caption{Number of augmentations versus classifier output on a test example. Lower is better. Performance plateaus at around 25 augmentations, in this example.}
    \label{fig:num_augs}
\end{figure}

For each number of augmentations, we generated 3 augmented datasets (using different random seeds) and trained 3 models on each dataset (again using different random seeds), for a total of 9 data points for each number of augmentations. The results are shown in Figure \ref{fig:num_augs}. The y-axis is the classifier output on the test example, which has a true label of 0 (lower is better). Without any augmentations, the classifier has an output near 1, which is incorrect. Improvement began around 10 augmentations, and plateaued by 20. Since computation cost was not significant, we chose 25 augmentations per-example for all experiments in the main text.

\subsection{Hyperparameters}

The maximum number of iterations for stepping and projecting are $\nStep=5$ and $\mProj=25$, and we stop the outer loop if the change in transform is ever less than $\projectionNotProgressing=0.001$. When solving Problem 3, we stop if the gradient is smaller than $\projStepSizeThreshold=0.0003$. There are also learning rate, learning rate decay, and weighting parameters used when solving Problem 3. For the objective function weighting terms, we use $\betaBbox=0.05,\betaValid=1,\betaOcc=1,\betaDmd=0.1$. The weighting terms were tuned so that the magnitude of the different weighted losses terms were comparable, and learning rate and number of iterations were tuned to maximize convergence. All values used are documented in our code, which can be found on our \href{https://sites.google.com/view/data-augmentation4manipulation}{project website}.